\documentclass[10pt,twocolumn,letterpaper]{article}

\usepackage{cvpr}
\usepackage{times}
\usepackage{epsfig}
\usepackage{graphicx}
\usepackage{amsmath}
\usepackage{amssymb}
\usepackage{booktabs}
\usepackage{flushend}

\usepackage[ruled,linesnumbered]{algorithm2e}

\usepackage[pagebackref=true,breaklinks=true,letterpaper=true,colorlinks,bookmarks=false]{hyperref}

\cvprfinalcopy

\def\cvprPaperID{948}

\ifcvprfinal\fi

\begin{document}

\title{Neural Rejuvenation: Improving Deep Network Training by\\
Enhancing Computational Resource Utilization}

\author{
Siyuan Qiao$^{1}$\thanks{Work done while an intern at Adobe.}~~~~Zhe Lin$^{2}$~~~~Jianming Zhang$^{2}$~~~~Alan Yuille$^{1}$\\
$^{1}$Johns Hopkins University~~~~$^{2}$Adobe Research\\
{\tt\small \{siyuan.qiao, alan.yuille\}@jhu.edu~~~~\{zlin,jianmzha\}@adobe.com}
}

\maketitle
\thispagestyle{empty}

\begin{abstract}
In this paper, we study the problem of improving computational resource utilization of neural networks.
Deep neural networks are usually over-parameterized for their tasks in order to achieve good performances, thus are likely to have underutilized computational resources.
This observation motivates a lot of research topics, e.g.~network pruning, architecture search, etc.
As models with higher computational costs (e.g.~more parameters or more computations) usually have better performances, we study the problem of improving the resource utilization of neural networks so that their potentials can be further realized.
To this end, we propose a novel optimization method named Neural Rejuvenation.
As its name suggests, our method detects dead neurons and computes resource utilization in real time, rejuvenates dead neurons by resource reallocation and reinitialization, and trains them with new training schemes.
By simply replacing standard optimizers with Neural Rejuvenation, we are able to improve the performances of neural networks by a very large margin while using similar training efforts and maintaining their original resource usages.
\end{abstract}

\section{Introduction}
Deep networks achieve state-of-the-art performances in many visual tasks~\cite{deeplab,resnet,mrnn,fewshot}.
On large-scale tasks such as ImageNet~\cite{ILSVRC15} classification, a common observation is that the models with more parameters, or more FLOPs, tend to achieve better results.
For example, DenseNet~\cite{densenet} plots the validation error rates as functions of the number of parameters and FLOPs, and shows consistent accuracy improvements as the model size increases.
This is consistent with our intuition that large-scale tasks require models with sufficient capacity to fit the data well.
As a result, it is usually beneficial to train a larger model if the additional computational resources are properly utilized.
However, previous work on network pruning~\cite{netslim,bnprune} already shows that many neural networks trained by SGD have unsatisfactory resource utilization.
For instance, the number of parameters of a VGG~\cite{vggnet} network trained on CIFAR~\cite{cifar} can be compressed by a factor of 10 without affecting its accuracy~\cite{netslim}.
Such low utilization results in a waste of training and testing time, and restricts the models from achieving their full potentials.
To address this problem, we investigate novel neural network training and optimization techniques to enhance resource utilization and improve accuracy.

Formally, this paper studies the following optimization problem.
We are given a loss function $\mathcal{L}(f(x;\mathcal{A},\theta_{\mathcal{A}}), y)$ defined on data $(x,y)$ from a dataset $\mathcal{D}$, and a computational resource constraint $\mathcal{C}$.
Here, $f(x;\mathcal{A}, \theta_{\mathcal{A}})$ is a neural network with architecture $\mathcal{A}$ and parameterized by $\theta_{\mathcal{A}}$.
Let $c(\mathcal{A})$ denote the cost of using architecture $\mathcal{A}$ in $f$, \textit{e.g.},
$c(\mathcal{A})$ can be the number of parameters in $\mathcal{A}$ or its FLOPs.
Our task is to find $\mathcal{A}$ and its parameter $\theta_{\mathcal{A}}$ that minimize the average loss $\mathcal{L}$ on dataset $\mathcal{D}$ under the resource constraint $\mathcal{C}$, \textit{i.e.},
\begin{align}\label{eq:1}
\begin{split}
    \mathcal{A},\theta_{\mathcal{A}} =& \arg\min_{\mathcal{A},\theta_{\mathcal{A}}}\frac{1}{N}\sum_{i=1}^N\mathcal{L}\big(f(x_i;\mathcal{A},\theta_{\mathcal{A}}), y_i\big)\\
    & \text{\it s.t.}~~c(\mathcal{A})\leq\mathcal{C}
\end{split}
\end{align}

The architecture $\mathcal{A}$ is usually designed by researchers and fixed during minimizing Eq.~\ref{eq:1}, and thus the solution $\mathcal{A},\theta_{\mathcal{A}}$ will always meet the resource constraint.
When $\mathcal{A}$ is fixed, $\theta_\mathcal{A}$ found by standard gradient-based optimizers may have neurons (\textit{i.e.}~channels) that have little effects on the average loss, removing which will save resources while maintaining good performance.
In other words, $\theta_{\mathcal{A}}$ may not fully utilize all the resources available in $\mathcal{A}$.
Let $\mathcal{U}(\theta_\mathcal{A})$ denote the computational cost based on $\theta_{\mathcal{A}}$'s actual utilization of the computational resource of $\mathcal{A}$, which can be measured by removing dead neurons which have little effect on the output.
Clearly, $\mathcal{U}(\theta_{\mathcal{A}})\leq c(\mathcal{A})$.
As previous work suggests~\cite{netslim}, the utilization ratio $r(\theta_{\mathcal{A}})=\mathcal{U}(\theta_{\mathcal{A}}) / c(\mathcal{A})$ trained by standard SGD can be as low as $11.5\%$.

The low utilization motivates the research on network pruning~\cite{netslim,bnprune}, \textit{i.e.}, extracting the effective subnet $\mathcal{A'}$ from $\mathcal{A}$ such that $c(\theta_{\mathcal{A}'})=\mathcal{U}(\theta_\mathcal{A})$.
Although the utilization ratio $r(\theta_{\mathcal{A}'})$ is high, this is opposite to our problem because it tries to narrow the difference between $c(\mathcal{\mathcal{A}})$ and $\mathcal{U}(\mathcal{\mathcal{A}})$ by moving $c(\mathcal{\mathcal{A}})$ towards $\mathcal{U}(\mathcal{\mathcal{A}})$.
By contrast, our objective is to design an optimization procedure $\mathcal{P}$ which enables us to find parameters $\theta_{\mathcal{A}}=\mathcal{P}(\mathcal{A}, \mathcal{L}, \mathcal{D})$ with a high $r(\theta_\mathcal{A})$.
In other words, we are trying to move $\mathcal{U}(\mathcal{\mathcal{A}})$ towards $c(\mathcal{\mathcal{A}})$, which maximizes the real utilization of the constraint $\mathcal{C}$.

There are many reasons for low utilization ratio $r(\theta_{\mathcal{A}})$.
One is bad initialization~\cite{frankle2018lottery},
which can be alleviated by parameter reinitialization for the spare resource.
Another one is inefficient resource allocation~\cite{morphnet}, \textit{e.g.}, the numbers of channels or the depths of blocks may not be configured properly to meet their real needs.
Unlike the previous methods \cite{morphnet,pnas} which search architectures by training a lot of networks, we aim to design an optimizer that trains \textit{one} network only \textit{once} and includes both resource \textit{reinitialization} and \textit{reallocation} for maximizing resource utilization.

In this paper, we propose an optimization method named Neural Rejuvenation (NR) for enhancing resource utilization during training.
Our method is intuitive and simple.
During training, as some neurons may be found to be useless (\textit{i.e.}~have little effect on the output), we revive them with new initialization and allocate them to the places they are needed the most.
From a neuroscience perspective, this is to rejuvenate dead neurons by bringing them back to functional use~\cite{NR} -- hence the name.
The challenges of Neural Rejuvenation are also clear.
Firstly, we need a real-time resource utilization monitor.
Secondly, when we rejuvenate dead neurons, we need to know how to reinitialize them and where to place them.
Lastly, after dead neuron rejuvenation, survived neurons ($\mathcal{S}$ neurons) and rejuvenated neurons ($\mathcal{R}$ neurons) are mixed up, and how to train networks with both of them present is unclear.

Our solution is a plug-and-play optimizer, the codes of which will be made public.
Under the hood, it is built on standard gradient-based optimizers, but with additional functions including real-time resource utilization monitoring, dead neuron rejuvenation, and new training schemes designed for networks with mixed types of neurons.
We introduce these components as below.

\vspace{-0.2in}
\paragraph{Resource utilization monitoring}
Similar to \cite{netslim,bnprune}, we use the activation scales of neurons to identify utilized and spare computational resource, and calculate a real-time utilization ratio $r(\theta_{\mathcal{A}})$ during training.
An event will be triggered if $r(\theta_{\mathcal{A}})$ is below a threshold $T_r$, and the procedure of dead neuron rejuvenation will take the control before the next step of training, after which $r(\theta_{\mathcal{A}})$ will go back to $1$.

\vspace{-0.2in}
\paragraph{Dead neuron rejuvenation}
This component rejuvenates the dead neurons by collecting the unused resources and putting them back in $\mathcal{A}$.
Similar to MorphNet~\cite{morphnet}, more spare resources are allocated to the layers with more $\mathcal{S}$ neurons.
However, unlike MorphNet~\cite{morphnet} which trains the whole network again from scratch after the rearrangement, we only reinitialize the dead neurons and then continue training.
By taking the advantages of dead neuron reinitialization~\cite{frankle2018lottery} and our training schemes, our optimizer is able to train \textit{one} model only \textit{once} and outperform the optimal network found by MorphNet~\cite{morphnet} from lots of architectures.

\vspace{-0.2in}
\paragraph{Training with mixed neural types}
After dead neuron rejuvenation, each layer will have two types of neurons: $\mathcal{S}$ and $\mathcal{R}$ neurons.
We propose two novel training schemes for different cases when training networks with mixed types of neurons.
The first one is to remove the cross-connections between $\mathcal{S}$ and $\mathcal{R}$ neurons, and the second one is to use cross-attention between them to increase the network capacity.
Sec.~\ref{sec:lat} presents the detailed discussions.

\vspace{0.05in}
We evaluate Neural Rejuvenation on two common image recognition benchmarks, \textit{i.e.} CIFAR-10/100~\cite{cifar} and ImageNet~\cite{ILSVRC15} and show that it outperforms the baseline optimizer by a very large margin.
For example, we lower the top-1 error of ResNet-50~\cite{resnet} on ImageNet by 1.51\%, and by 1.82\% for MobileNet-0.25~\cite{howard2017mobilenets} while maintaining their FLOPs.
On CIFAR where we rejuvenate the resources to the half of the constraint and compare with the previous state-of-the-art compression method~\cite{netslim}, we outperform it by up to 0.87\% on CIFAR-10 and 3.39\% on CIFAR-100.

\section{Related Work}\label{sec:related}
\paragraph{Efficiency of neural networks}
It is widely recognized that deep neural networks are over-parameterized~\cite{ba2014deep,denton2014exploiting} to win the filter lottery tickets~\cite{frankle2018lottery}.
This efficiency issue is addressed by many methods, including weight quantization~\cite{courbariaux2016binarized,rastegari2016xnor}, low-rank approximation~\cite{denton2014exploiting,lebedev2014speeding}, knowledge distillation~\cite{hinton2015distilling,yang2018knowledge} and network pruning~\cite{han2015learning,hassibi1993second,lecun1990optimal,li2016pruning,netslim,molchanov2016pruning,bnprune,yu2017nisp}.
The most related method is network pruning, which finds the subnet that affects the outputs the most.
Network pruning has several research directions, such as weight pruning, structural pruning, \textit{etc}.
Weight pruning focuses on individual weights~\cite{han2015deep,han2015learning,hassibi1993second,lecun1990optimal}, but requires dedicated hardware and software implementations to achieve compression and acceleration~\cite{han2016eie}.
Structural pruning identifies channels and layers to remove from the architecture, thus is able to directly achieve speedup without the need of specialized implementations~\cite{alvarez2016learning,chnprune,lebedev2016fast,luo2017thinet,molchanov2016pruning,wen2016learning,zhou2016less}.
Following~\cite{netslim,bnprune}, we encourage channel sparsity by imposing penalty term to the scaling factors.

Different from these previous methods, Neural Rejuvenation studies the efficiency issue from a new angle: we aim to directly maximize the utilization by reusing spare computational resources.
As an analogy in the context of lottery hypothesis~\cite{frankle2018lottery}, Neural Rejuvenation is like getting refund for the useless tickets and then buying new ones.

\vspace{-0.15in}
\paragraph{Cross attention}
In this work, we propose to use cross attention to increase the capacity of the networks without introducing additional costs.
This is motivated by adding second-order transform~\cite{goggin1991second,kazemy2007second,sort} on multi-branch networks~\cite{resnet,densenet,gunn,srivastava2015training,szegedy2015going,mssar}.
Instead of using a geometric mean as in~\cite{sort}, we propose to use cross attention~\cite{hao2017end,lee2018stacked} as the second-order term to increase the capacity.
Attention models have been widely used in deep neural networks for a variety of vision and language tasks, such as object detection~\cite{ba2014multiple,mnih2014recurrent,scalenet,des}, machine translation~\cite{bahdanau2014neural}, visual question answering~\cite{chen2015abc,xu2016ask}, image captioning~\cite{xu2015show}, \textit{etc}.
Unlike the previous attention models, our method uses one group of channels to generate attentions for the other channels, and our attention model is mainly used to increase capacity.

\vspace{-0.15in}
\paragraph{Architecture search}
Our objective formulated by Eq.~\ref{eq:1} is similar to neural architecture search which approaches the problem by searching architecture $\mathcal{A}$ in a pre-defined space, and thus they need to train a lot of networks to find the optimal architecture.
For example, NAS~\cite{zoph2016neural} uses reinforcement learning to find the architecture, \cite{zoph2017learning} extends it by using a more structured search space, and \cite{pnas} improves the search efficiency by progressively finding architectures.
But their computational costs are very high, \textit{e.g.}, \cite{zoph2017learning} uses 2000 GPU days.
There are more methods focusing on the search problem~\cite{baker2016designing,cai2018efficient,elsken2017simple,miikkulainen2019evolving,pham2018efficient,real2017large,zhong2017practical}.
Different from architecture search, Neural Rejuvenation does not search $\mathcal{A}$ which requires hundreds of thousands of models to train, although it does change the architecture a little bit.
Instead, our method is an optimization technique which trains models in just one training pass.
The closest method is MorphNet~\cite{morphnet} in that we both use linearly expanding technique to find resource arrangement. Yet, it still needs multiple training passes and does not rejuvenate dead neurons nor reuse partially-trained filters.
We show direct comparisons with it and outperform it by a large margin.

\vspace{-0.15in}
\paragraph{Parameter reinitialization}
Parameter reinitialization is a common strategy in optimization to avoid useless computations and improve performances.
For example,  during the k-means optimization, empty clusters are automatically reassigned, and big clusters are encouraged to split into small clusters~\cite{caron2018deep,johnson2017billion,joulin2016learning,xu2005maximum}.
Our method is reminiscent to this in that it also detects unsatisfactory components and reinitializes them so that they can better fit the tasks.

\section{Neural Rejuvenation}

Algorithm~\ref{alg:nr} presents a basic framework of Neural Rejuvenation which adds two new modules: resource utilization monitoring (Step 6) and dead neuron rejuvenation (Step 7 and 8) to a standard SGD optimizer.
The training schemes are not shown here, which will be discussed in Sec.~\ref{sec:lat}.
We periodically set the Neural Rejuvenation flag on with a pre-defined time interval to check the utilization and rejuvenate dead neurons when needed.
In the following subsections, we will present how each component is implemented.

\subsection{Resource Utilization Monitoring}
\subsubsection{Liveliness of Neurons}
We consider a convolutional neural network where every convolutional layer is followed by a batch normalization layer~\cite{batchnorm}.
An affine transform layer with learnable parameters are also valid if batch normalization is not practical.
For each batch-normalized convolutional layer, let $\mathcal{B}=\{u_1, ..., u_m\}$ be a mini-batch of values after the convolution.
Then, its normalized output $\{v_1, ..., v_m\}$ is
\begin{align}
\begin{split}
    & v_i = \gamma\cdot \frac{u_i-\mu_{\mathcal{B}}}{\sqrt{\sigma_{\mathcal{B}}^2+\epsilon}} + \beta,~~\forall i\in\{1,...,m\} \\
    \text{where}&~\mu_{\mathcal{B}} =~\frac{1}{m}\sum_{i=1}^{m}u_i ~\text{and}~
    \sigma^2_{\mathcal{B}}=\frac{1}{m}\sum_{i=1}^m(u_i-\mu_{\mathcal{B}})^2
\end{split}
\end{align}
Each neuron (\textit{i.e.} channel) in the convolutional layer has its own learnable scaling parameter $\gamma$, which we use as an estimate of the liveliness of the corresponding neuron~\cite{netslim,bnprune}.
As our experiments suggest, if a channel's scaling parameter $\gamma$ is less than $0.01\times\gamma_{\max}$ where $\gamma_{\max}$ is the maximum $\gamma$ in the same batch-normalized convolution layer, removing it will have little effect on the output of $f$ and the loss $\mathcal{L}$.
Therefore, in all experiments shown in this paper, a neuron is considered dead if its scaling parameter $\gamma<0.01\times\gamma_{\max}$.
Let $\mathcal{T}$ be the set of all the scaling parameters within the architecture $\mathcal{A}$.
Similar to \cite{netslim}, we add a L1 penalty term on $\mathcal{T}$ in order to encourage neuron sparsity, \textit{i.e.,}~instead of the given loss function $\mathcal{L}$, we minimize the following loss
\begin{equation}
    \mathcal{L}_\lambda = \mathcal{L}\big(f(x_i;\mathcal{A},\theta_{\mathcal{A}}), y_i\big) + \lambda \sum_{\gamma\in\mathcal{T}} |\gamma|
\end{equation}
where $\lambda$ is a hyper-parameter.

\begin{algorithm}[t]
\small
\SetStartEndCondition{ }{}{}%
\SetKwProg{Fn}{def}{\string:}{}
\SetKwFunction{Range}{range}%%
\SetKw{KwTo}{in}\SetKwFor{For}{for}{\string:}{}%
\SetKwIF{If}{ElseIf}{Else}{if}{:}{elif}{else:}{}%
\SetKwFor{While}{while}{:}{fintq}%
\AlgoDontDisplayBlockMarkers\SetAlgoNoEnd%
\SetKwInOut{Input}{Input}
\SetKwInOut{Output}{Output}
\caption{SGD with Neural Rejuvenation}\label{alg:nr}
\Input{Learning rate $\epsilon$, utilization threshold $T_r$, initial architecture $\mathcal{A}$ and $\theta_{\mathcal{A}}$, and resource constraint $\mathcal{C}$}
\While{stopping criterion not met}{
    Sample a minibatch $\{(x_1, y_1), ..., (x_m, y_m)\}$\;

    Compute gradient $g\leftarrow \frac{1}{m}\nabla\sum_i\mathcal{L}(f(x_i;\mathcal{A},\theta_\mathcal{A}), y_i)$\;

    Apply update $\theta_\mathcal{A} = \theta_\mathcal{A} - \epsilon\cdot g$\;

    \If{neural rejuvenation flag is on}{
        Compute utilization ratio $r(\theta_{\mathcal{A}})$\;

        \If{$r(\theta_{\mathcal{A}})<T_r$}{
            Rejuvenate dead neurons and obtain new $\mathcal{A}$ and $\theta_{\mathcal{A}}$ under resource constraint $\mathcal{C}$\;

        }
    }
}
\Return Architecture $\mathcal{A}$ and its parameter $\theta_{\mathcal{A}}$\;
\end{algorithm}

\subsubsection{Computing $r(\theta_{\mathcal{A}})$ by Feed-Forwarding}
Here, we show how to compute the utilization ratio $r(\theta_{\mathcal{A}})$ based on the liveliness of the neurons in real time.
We compute $r(\theta_{\mathcal{A}})$ by a separate feed-forwarding similar to that of function $f$.
The computational cost of the feed-forwarding for $r(\theta_{\mathcal{A}})$ is negligible compared with that of $f$.
We first rewrite the function $f$:
\begin{equation}
    f(x) = (f_l\circ f_{l-1}\circ ... \circ f_1) (x)
\end{equation}
where $f_i$ is the i-th layer of the architecture $\mathcal{A}$.
When computing $r(\theta_\mathcal{A})$, instead of passing the output of a layer computed from $x$ to the next layer as input, each layer $f_i$ will send a binary mask indicating the liveliness of its neurons.
Let $M_i^{\text{in}}$ denote the binary mask for the input neurons for layer $f_i$, and $M_i^{\text{out}}$ denote the binary mask for its own neurons.
Then, the effective number of parameters of $f_i$ is $||M_i^{\text{in}}||_1\cdot ||M_i^{\text{out}}||_1\cdot K_w\cdot K_h$, if $f_i$ is a convolutional layer with 1 group and no bias, and its computational cost is computed by $||M_i^{\text{in}}||_1\cdot ||M_i^{\text{out}}||_1\cdot K_w\cdot K_h\cdot O_w\cdot O_h$ following \cite{resnet}.
Here, $K_w$ and $K_h$ are the kernel size, and $O_w$ and $O_h$ are the output size.
Note that the cost of $f$ is the sum of the costs of all layers $f_i$; therefore, we also pass the effective computational cost and the original cost in feed-forwarding.
After that, we are able to compute $U(\theta_{\mathcal{A}})$ and $c(\theta_{\mathcal{A}})$, and consequently $r(\theta_{\mathcal{A}})$.
During the computation of $r(\theta_\mathcal{A})$, each layer will also keep a copy of the liveliness of the neurons of its previous layer.
This information is used in the step of dead neural rejuvenation after $r(\theta_\mathcal{A})<T_r$ is met.
It also records the values of the scaling parameter $\gamma$ of the input neurons.
This is used for neural rescaling which is discussed in Sec.~\ref{sec:neural_rescaling}.

\subsubsection{Adaptive Penalty Coefficient $\lambda$}
The utilization ratio $r(\theta_{\mathcal{A}})$ will depend on the value of the sparsity coefficient $\lambda$ as a larger $\lambda$ tends to result in a sparser network.
When $\lambda=0$, all neurons will probably stay alive as we have a tough threshold $0.01\times\gamma_{\max}$.
As a result, Step 7 and 8 of Algorithm~\ref{alg:nr} will never get executed and our optimizer is behaving as the standard one.
When $\lambda$ goes larger, the real loss function $\mathcal{L}_\lambda$ we optimize will become far from the original loss $\mathcal{L}$.
Consequently, the performance will be less unsatisfactory.
Therefore, choosing a proper $\lambda$ is critical for our problem, and we would like it to be automatic and optimized to the task and the architecture.

In Neural Rejuvenation, the value of $\lambda$ is dynamically determined by the trend of the utilization ratio $r(\theta_\mathcal{A})$.
Specifically, when the neural rejuvenation flag is on, we keep a record of the utilization ratio $r(\theta_\mathcal{A})^{t}$ after training for $t$ iterations.
After $\Delta t$ iterations , we compare the current ratio $r(\theta_\mathcal{A})^t$ with the previous one $r(\theta_\mathcal{A})^{t-\Delta t}$.
If $r(\theta_\mathcal{A})^t < r(\theta_\mathcal{A})^{t-\Delta t} - \Delta r$, we keep the current $\lambda$; otherwise, we increase $\lambda$ by $\Delta\lambda$.
Here, $\Delta t$, $\Delta r$ and $\Delta \lambda$ are hyper-parameters.
$\lambda$ is initialized with $0$.
After Step 8 gets executed, $\lambda$ is set back to $0$.

It is beneficial to set $\lambda$ in the above way rather than having a fixed value throughout the training.
Firstly, different tasks and architectures may require different values of $\lambda$. The above strategy frees us from manually selecting one based on trial and error.
Secondly, the number of iterations needed to enter Step 8 is bounded.
This is because after $\lambda$ gets large enough, each $\Delta t$ will decrease the utilization ratio by at least $\Delta r$.
Hence, the number of iterations to reach $T_{r}$ is bounded by $(1-T_{r}) / \Delta r + O(1)$.
In a word, this strategy automatically finds the value of $\lambda$, and guarantees that the condition $r(\theta_\mathcal{A}) < T_r$ will be met in a bounded number of training iterations.

\subsection{Dead Neuron Rejuvenation}
After detecting the liveliness of the neurons and the condition $r(\theta_{\mathcal{A}}) < T_r$ is met, we proceed to Step 8 of Algorithm~\ref{alg:nr}.
Here, our objective is to rejuvenate the dead neurons and reallocate those rejuvenated neurons to the places they are needed the most under the resource constraint $\mathcal{C}$.
There are three major steps in dead neuron rejuvenation.
We present them in order as follows.

\vspace{-0.15in}
\paragraph{Resource reallocation}\label{sec:neural_rescaling}
The first step is to reallocate the computational resource saved by removing all the dead neurons.
The removal reduces the computational cost from $c(\mathcal{A})$ to $U(\mathcal{\theta_{\mathcal{A}}})$; therefore, there is $c(\mathcal{A})-U(\mathcal{\theta_{\mathcal{A}}})$ available resource to reallocate.
The main question is where to add this free resource back in $\mathcal{A}$.
Let $w_i$ denote the number of output channels of layer $f_i$ in $f$, and $w_i$ is reduced to $w_i'$ by dead neuron removal.
Let $\mathcal{A}'$ denote the architecture after dead neuron removal with $w_i'$ output channels at layer $f_i$.
Then, $c(\mathcal{A'})=U(\mathcal{A})$.
To increase the computational cost of $\mathcal{A'}$ to the level of $\mathcal{A}$,
our resource reallocation will linearly expand $w_i''=\alpha\cdot w_i'$ by a shared expansion rate $\alpha$ across all the layers $f_i$, to build a new architecture $\mathcal{A}''$ with numbers of channels $w_i''$.
The assumption here is that if a layer has a higher ratio of living neurons, this layer needs more resources, \textit{i.e.}~more output channels;
by contrast, if a layer has a lower ratio, this means that more than needed resources were allocated to it in $\mathcal{A}$.
This assumption is modeled by having a shared linear expansion rate $\alpha$.

The resource reallocation used here is similar to the iterative squeeze-and-expand algorithm in MorphNet~\cite{morphnet} for neural architecture search.
The differences are also clear.
Neural Rejuvenation models both dead neuron reinitialization, reallocation and training schemes to train just one network only once, while MorphNet is only interested in the numbers of channels of each layer that are optimal when trained from scratch and finds it by training many networks.

\vspace{-0.15in}
\paragraph{Parameter reinitialization}
The second step is to reinitialize the parameters of the reallocated neurons.
Let $\mathcal{S}_{\text{in}}$ and $\mathcal{R}_{\text{in}}$ denote the input $S$ (survived) neurons and $R$ (rejuvenated) neurons, respectively, and $\mathcal{S}_{\text{out}}$ and $\mathcal{R}_{\text{out}}$ denote the output $S$ neurons and $R$ neurons, respectively.
Then, the parameters $W$ can be divided into four groups:
$W_{\mathcal{S}\rightarrow\mathcal{S}}$, $W_{\mathcal{S}\rightarrow\mathcal{R}}$,
$W_{\mathcal{R}\rightarrow\mathcal{R}}$,
$W_{\mathcal{R}\rightarrow\mathcal{S}}$,
which correspond to the parameters from $\mathcal{S}_{\text{in}}$ to $\mathcal{S}_{\text{out}}$, from $\mathcal{S}_{\text{in}}$ to $\mathcal{R}_{\text{out}}$, from $\mathcal{R}_{\text{in}}$ to $\mathcal{R}_{\text{out}}$ and from $\mathcal{R}_{\text{in}}$ to $\mathcal{S}_{\text{out}}$, respectively.
During reinitialization, the parameters $W_{\mathcal{S}\rightarrow\mathcal{S}}$ are kept since they survive the dead neuron test.
The parameters $W_{\mathcal{R}\rightarrow\mathcal{R}}$ are randomly initialized and their scaling parameters $\gamma$'s are restored to the initial level.
In order for the $\mathcal{S}$ neurons to keep their mapping functions after the rejuvenation, $W_{\mathcal{R}\rightarrow\mathcal{S}}$ is set to $0$.
We also set $W_{\mathcal{S}\rightarrow\mathcal{R}}$ to $0$ as this initialization does not affect the performances as the experiments suggest.

\vspace{-0.15in}
\paragraph{Neural rescaling}
Recall that in order to encourage the sparsity of the neurons, all neurons receive the same amount of penalty.
This means that not only the dead neurons have small scaling values, some $\mathcal{S}$ neurons also have scaling values that are very small compared with $\gamma_{\max}$ of the corresponding layers.
As experiments in Sec.~\ref{sec:abl} show, this is harmful for gradient-based training.
Our solution is to rescale those neurons to the initial level, \textit{i.e.}, $|\gamma_i'| = \max\{|\gamma_i|, \gamma_0\}~\forall i$, where $\gamma_0$ is the initial value for $\gamma$.
We do not change the sign of $\gamma$.
Note that rescaling takes all neurons into consideration, including $\mathcal{S}$ neurons with large scaling values ($|\gamma|\geq|\gamma_0|$), $\mathcal{S}$ neurons with small scaling values ($|\gamma|<|\gamma_0|$) and dead neurons ($|\gamma|\approx 0$).
After neural rescaling, we adjust the parameters to restore the original mappings.
For $\mathcal{S}$ neurons, let $s_i=\gamma_i'/\gamma_i$.
In order for $\mathcal{S}$ neurons to keep their original mapping functions, we divide the parameters that use them by $s_i$.
Experiments show that this leads to performance improvements.

\subsection{Training with Mixed Types of Neurons}\label{sec:lat}
Let us now focus on each individual layer.
After neural rejuvenation, each layer will have two types of input neurons, $\mathcal{S}_{\text{in}}$ and $\mathcal{R}_{\text{in}}$, and two types of output neurons, $\mathcal{S}_{\text{out}}$ and $\mathcal{R}_{\text{out}}$.
For simplicity, we also use them to denote the features of the corresponding neurons.
Then, by the definition of convolution, we have
\begin{align}\label{eq:normal}
\begin{split}
    \mathcal{S}_{\text{out}} &= W_{\mathcal{S}\rightarrow\mathcal{S}}*\mathcal{S}_{\text{in}} + W_{\mathcal{R}\rightarrow\mathcal{S}}*\mathcal{R}_{\text{in}} \\
    \mathcal{R}_{\text{out}} &= W_{\mathcal{S}\rightarrow\mathcal{R}}*\mathcal{S}_{\text{in}} + W_{\mathcal{R}\rightarrow\mathcal{R}}*\mathcal{R}_{\text{in}}
\end{split}
\end{align}
where $*$ denote the convolution operation.
$W_{\mathcal{R}\rightarrow\mathcal{S}}$ is set to $0$ in reinitialization;
therefore, $\mathcal{S}_{\text{out}} = W_{\mathcal{S}\rightarrow\mathcal{S}}*\mathcal{S}_{\text{in}}$ initially, which keeps the original mappings between $\mathcal{S}_{\text{in}}$ and $\mathcal{S}_{\text{out}}$.
In this subsection, we discuss how to train $W$.

The training of $W$ depends on how much the network needs the additional capacity brought by the rejuvenated neurons to fit the data.
When $\mathcal{S}$ neurons do not need this additional capacity at all, adding $\mathcal{R}$ neurons by Eq.~\ref{eq:normal} may not help because $\mathcal{S}$ neurons alone are already able to fit the data well.
As a result, changing training scheme is necessary in this case in order to utilize the additional capacity.
However, when $\mathcal{S}$ neurons alone have difficulties fitting the data, the additional capacity provided by $\mathcal{R}$ neurons will ease the training.
They were found to be useless previously either because of improper initialization or inefficient resource arrangement, but now are reinitialized and rearranged.
We present the detailed discussions as below.

\vspace{-0.15in}
\paragraph{When $\mathcal{S}$ does not need $\mathcal{R}$}
Here, we consider the situation where the network capacity is bigger than necessary, and $\mathcal{S}$ neurons alone are able to fit the training data well.
An example is training networks on CIFAR~\cite{cifar}, where most of the modern architectures can reach $99.0\%$ training accuracy.
When adding $\mathcal{R}$ neurons into the architecture as in Eq.~\ref{eq:normal}, since $\mathcal{S}$ neurons have already been trained to fit the data well, the gradient back-propagated from the loss will not encourage any great changes on the local mapping $(\mathcal{S}_{\text{in}}, \mathcal{R}_{\text{in}}) \rightarrow(\mathcal{S}_{\text{out}}, \mathcal{R}_{\text{out}})$.
Therefore, keep modeling the computation as Eq.~\ref{eq:normal} may result in $\mathcal{R}_{\text{in}}$ neurons being dead soon and $\mathcal{R}_{\text{out}}$ producing redundant features.

The cause of the above problem is the existence of cross-connections between $\mathcal{R}$ neurons and $\mathcal{S}$ neurons, which provides short-cuts to $\mathcal{R}$.
If we completely remove them, \textit{i.e.},
\begin{align}\label{eq:no}
\begin{split}
    \mathcal{S}_{\text{out}} = W_{\mathcal{S}\rightarrow\mathcal{S}}*\mathcal{S}_{\text{in}}~~~
    \mathcal{R}_{\text{out}} = W_{\mathcal{R}\rightarrow\mathcal{R}}*\mathcal{R}_{\text{in}}
\end{split}
\end{align}
then $\mathcal{R}$ neurons are forced to learn features that are new and ideally different.
We use NR-CR to denote Neural Rejuvenation with cross-connections removed.

\vspace{-0.15in}
\paragraph{When $\mathcal{S}$ needs $\mathcal{R}$}
Here, we assume that the capacity of $\mathcal{S}$ alone is not enough for fitting the training data.
One example is training small networks on ImageNet dataset~\cite{ILSVRC15}.
In this case, it is desirable to keep the cross-connections to increase the capacity.
Experiments in Sec.~\ref{sec:abl} compare the performances of a simplified VGG network~\cite{vggnet} on ImageNet, and show that Neural Rejuvenation with cross-connections kept and removed both improve the accuracies, but keeping cross-connections improves more.

\vspace{-0.15in}
\paragraph{Cross-attention between $\mathcal{S}$ and  $\mathcal{R}$}
We continue the discussion where we assume $\mathcal{S}$ needs the capacity of $\mathcal{R}$ and we keep the cross-connections.
Then according to Eq.~\ref{eq:normal}, the outputs from $\mathcal{S}_{\text{in}}$ and $\mathcal{R}_{\text{in}}$ are added up for $\mathcal{S}_{\text{out}}$, \textit{i.e.}
\begin{equation}\label{eq:ori}
    \mathcal{S}_{\text{out}} = W_{\mathcal{S}\rightarrow\mathcal{S}}*\mathcal{S}_{\text{in}} + W_{\mathcal{R}\rightarrow\mathcal{S}}*\mathcal{R}_{\text{in}}
\end{equation}
Since the assumption here is that the model capacity is insufficient for fitting the training data, it would be better if we can increase the capacity not only by rejuvenating dead neurons, but also by changing Eq.~\ref{eq:ori} to add more capacity without using any more parameters nor resulting in substantial increases of computations (if any) compared with the convolution operation itself.
As $W_{\mathcal{S}\rightarrow\mathcal{S}}*\mathcal{S}_{\text{in}}$ is fixed, we focus on $W_{\mathcal{R}\rightarrow\mathcal{S}}*\mathcal{R}_{\text{in}}$.
One way to increase capacity is to use second-order response transform~\cite{sort}.
The original second-order response transform is defined on residual learning~\cite{resnet} by adding a geometric mean, \textit{i.e.}
\begin{equation}
    y = x + F(x) \Rightarrow y = x + F(x) + \sqrt{x\cdot F(x)}
\end{equation}
For our problem, although Eq.~\ref{eq:ori} does not have residual connections, the outputs $W_{\mathcal{S}\rightarrow\mathcal{S}}*\mathcal{S}_{\text{in}}$ and
$W_{\mathcal{R}\rightarrow\mathcal{S}}*\mathcal{R}_{\text{in}}$ are added up as in residual learning; therefore, we can add a similar response transform to Eq.~\ref{eq:ori}.
Instead of adding a geometric mean which causes training instability~\cite{sort}, we propose to use cross attentions as shown in Eq.~\ref{eq:ca}.
\begin{equation}\label{eq:ca}
    \mathcal{S}_{\text{out}} = W_{\mathcal{S}\rightarrow\mathcal{S}}*\mathcal{S}_{\text{in}} + 2\cdot\sigma(W_{\mathcal{S}\rightarrow\mathcal{S}}*\mathcal{S}_{\text{in}})W_{\mathcal{R}\rightarrow\mathcal{S}}*\mathcal{R}_{\text{in}}
\end{equation}
Here, $\sigma(\cdot)$ denotes the Sigmoid function.
Symmetrically, we add cross attentions to the output of $\mathcal{R}_{\text{out}}$, \textit{i.e.}
\begin{equation}\label{eq:ca2}
    \mathcal{R}_{\text{out}} = W_{\mathcal{R}\rightarrow\mathcal{R}}*\mathcal{R}_{\text{in}} + 2\cdot\sigma(W_{\mathcal{R}\rightarrow\mathcal{R}}*\mathcal{R}_{\text{in}})W_{\mathcal{S}\rightarrow\mathcal{R}}*\mathcal{S}_{\text{in}}
\end{equation}
We use NR-CA to denote NR with cross attentions.

\section{Experiments}
In this section, we will show the experimental results that support our previous discussions, and present the improvements of Neural Rejuvenation on a variety of architectures.

\subsection{Resource Utilization}\label{sec:res_utl}

\begin{figure}[h]
    \centering
    \includegraphics[width=0.9\linewidth]{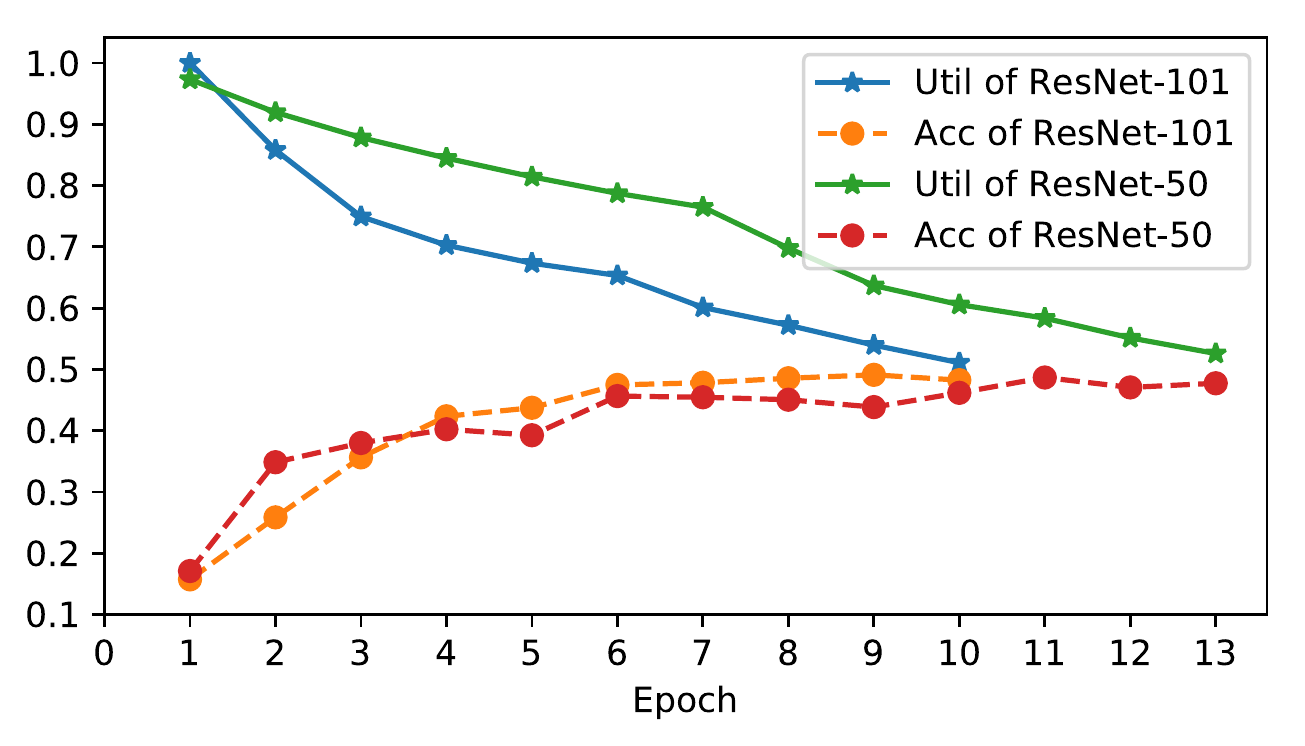}
    \caption{Parameter utilization and validation accuracy of ResNet-50 and ResNet-101 trained on ImageNet from scratch.}
    \label{fig:util_acc}
\end{figure}

We show the resource utilization of training ResNet-50 and ResNet-101 on ImageNet in Figure~\ref{fig:util_acc} when the sparsity term is added to the loss.
In the figure, we show the plots of the parameter utilization and validation accuracy of the models with respect to the number of training epochs.
Training such a model usually takes 90 epochs when the batch size is 256 or 100 epochs when the batch size is 128~\cite{densenet}.
In all the experiments, the sparsity coefficient $\lambda$ is initialized with 0, $\Delta t$ is set to one epoch, $\Delta r=0.01$ and $\Delta\lambda=5\times 10^{-5}$.
$T_r$ is set to $0.5$ unless otherwise stated.

Fig.~\ref{fig:util_acc} shows two typical examples that convey the following important messages.
(1) Training on large-scale dataset such as ImageNet cannot avoid the waste of the computational resources; therefore, our work is also valid for large-scale training.
(2) It is easier to find dead neurons in larger models than in smaller models.
This is consistent with our intuition that larger models increase the capacity and the risk of more resource wastes.
(3) It does not take too long to reach the utilization threshold at 0.5.
10 epochs are enough for saving half of the resources for ResNet-101.

For ImageNet training, we set the neural rejuvenation flag on only for the first $30$ epochs where the learning rate is $0.1$.
Since it usually takes 10-20 epochs for $r(\theta_{\mathcal{A}})$ to reach $T_r=0.5$, there will be about 1 to 2 times that Step 8 in Algorithm~\ref{alg:nr} will get executed.
To simplify the experiments, we only do one time of neural rejuvenation on ImageNet and reset the epoch counter to 0 afterwards.
The training time with neural rejuvenation thus will be a little longer than the original training, but the increase will be less than $20\%$ and experiments show that it is definitely worth it.
For unlimited training time, Sec.~\ref{sec:multi} shows the performances on CIFAR with multiple times of Neural Rejuvenation.

\subsection{Ablation Study on Neural Rejuvenation}\label{sec:abl}

To provide better understandings of Neural Rejuvenation applied on training deep networks, we present an ablation study shown in Table~\ref{tab:abl}, which demonstrates the results of Neural Rejuvenation with different variations.

\begin{table}[h]
\small
    \centering
    \begin{tabular}{l|cc|l|cc}
    \toprule
    Method & Top-1 & Top-5 & Method & Top-1 & Top-5\\
    \midrule
    BL & 32.13 & 11.97 & BL-CA & 31.58 & 11.46\\
    NR-CR & 31.40 & 11.53 &  NR-FS & 31.26 & 11.37 \\
    NR & 30.74 & 10.94 & NR-BR & 30.31 & 10.67 \\
    NR-CA & 30.28 & 10.88 & NR-CA-BR & 29.98 & 10.58\\
    \bottomrule
    \end{tabular}
    \vspace{0.05in}
    \caption{Error rates of a simplified VGG-19 on ImageNet with $T_r=0.25$ while maintaining the number of parameters.
    BL: baseline. BL-CA: baseline with cross attentions.
    NR-CR: Neural Rejuvenation with cross-connections removed.
    NR-FS: training $\mathcal{A}$ found by NR from scratch.
    NR: Neural Rejuvenation with cross-connections.
    NR-BR: Neural Rejuvenation with neural rescaling.
    NR-CA: Neural Rejuvenation with cross attentions.
    NR-CA-BR: Neural Rejuvenation with cross attentions and neural rescaling.}
    \label{tab:abl}
\end{table}

\begin{table*}[t]
\small
    \setlength{\tabcolsep}{0.43em}
    \centering
    \begin{tabular}{l | cc|cc  | cc|cc | cc|cc | c}
        \toprule
         Architecture &  \multicolumn{4}{c|}{Baseline} & \multicolumn{4}{c|}{NR Params} & \multicolumn{4}{c|}{NR FLOPs} & Relative\\
         \cmidrule{2-13}
         & Params & FLOPs & Top-1 & Top-5 & Params & FLOPs & Top-1 & Top-5 &  Params & FLOPs & Top-1 & Top-5 & Gain\\
        \midrule
        DenseNet-121~\cite{densenet} & 7.92M & 2.83G & 25.32 & 7.88 & 8.22M & 3.13G & 24.50 & 7.49 & 7.28M & 2.73G & 24.78 & 7.56 & -3.24\%\\
        VGG-16~\cite{vggnet} & 37.7M & 15.3G & 24.26 & 7.32 & 36.4M & 23.5G & 23.11 & 6.69 & 21.5M & 15.3G & 23.71 & 7.01 & -4.74\% \\
        ResNet-18~\cite{resnet} & 11.7M & 1.81G & 30.30 & 10.7 & 11.9M & 2.16G & 28.86 & 9.93 & 9.09M & 1.73G & 29.73 & 10.5 & -4.75\% \\
        ResNet-34~\cite{resnet} & 21.8M & 3.66G & 26.61 & 8.68 &  21.9M & 3.77G &  25.77 & 8.10 & 20.4M & 3.56G & 25.45 & 8.04 & -4.35\%\\
        ResNet-50~\cite{resnet} & 25.6M & 4.08G & 24.30 & 7.19 & 26.4M & 3.90G & 22.93 & 6.47 & 26.9M & 3.99G & 22.79 & 6.56 & -6.21\%\\
        ResNet-101~\cite{resnet} & 44.5M & 7.80G & 22.44 & 6.21 & 46.6M & 6.96G & 21.22 & 5.76 &  50.2M & 7.51G & 20.98 & 5.69 & -6.50\% \\
        \bottomrule
    \end{tabular}
    \vspace{0.05in}
    \caption{Error rates of deep neural networks on ImageNet validation set trained with and without Neural Rejuvenation.
    Each neural network has three sets of top-1 and top-5 error rates, which are baseline, Neural Rejuvenation with the number of parameters as the resource constraint (NR Params), and Neural Rejuvenation with FLOPs as resource constraint (NR FLOPs).
    The last column \textit{Relative Gain} shows the best relative gain of top-1 error while maintaining either number of parameters or FLOPs.
    }
    \label{tab:imagenet}
\end{table*}

The network is a simplified VGG-19, which is trained on low-resolution images from ImageNet.
The image size for training and testing is 128x128.
We remove the last three fully-connected layers, and replace them with a global average pooling layer and one fully-connected layer.
The resulted model has only $20.5$M parameters.
To further accelerate training, we replace the first convolutional layer with that in ResNet~\cite{resnet}.
By applying all the changes, we can train one model with 4 Titan Xp GPUs in less than one day, which is fast enough for the purpose of ablation study.

Clearly, such a simplified model does not have sufficient capacity for fitting ImageNet.
As we have discussed in Sec.~\ref{sec:lat}, it is better to keep the cross connections for increasing the model capacity.
As also demonstrated here, NR-CR improves the top-1 accuracy by $0.7\%$ than the baseline, but is $0.7\%$ behind NR where cross-connections are kept.
We further show that cross attentions lower the top-1 error rates by roughly $0.5\%$, and neural rescaling further improves the accuracies.
In the following experiments on ImageNet, we use NR-CA-BR for all the methods.

\subsection{Results on ImageNet}
Table~\ref{tab:imagenet} shows the performance improvements on ImageNet dataset~\cite{ILSVRC15}.
ImageNet dataset is a large-scale image classification dataset, which contains about 1.28 million color images for training and 50,000 for validation.
Table~\ref{tab:imagenet} lists some modern architectures which achieve very strong accuracies on such a challenging task.
Previously, a lot of attention is paid to designing novel architectures that are more suitable for vision tasks.
Our results show that in addition to architecture design and search, the current optimization technique still has a lot of room to improve. Our work focuses only on the utilization issues, but already achieves strong performance improvements.

Here, we briefly introduce the setting of the experiments for easy reproduction.
All the models are trained with batch size 256 if the model can fit in the memory; otherwise, we set the batch size to 128.
In total, we train the models for 90 epochs when the batch size is 256, and for 100 epochs if the batch size is 128.
The learning rate is initialized as 0.1, and then divided by 10 at the 31$^{\text{st}}$, 61$^{\text{st}}$, and 91$^{\text{st}}$ epoch.

For our task, we make the following changes to those state-of-the-art architectures.
For VGG-16~\cite{vggnet}, we add batch normalization layers after each convolutional layer and remove the last three fully-connected layers.
After that, we add two convolutional layers that both output 4096 channels, in order to follow the original VGG-16 that has two fully-connected layers outputting the same amount of channels.
After these two convolutional layers, we add a global average pooling layer, and a fully-connected layer that transforms the 4096 channels to 1000 channels for image classification.
The resulted model has fewer number of parameters (138M to 37.7M), but with a much lower top-1 error rate (27 to 24.26).
All the VGG-16 layers receive the sparsity penalty.
For ResNet~\cite{resnet}, all the convolutional layers except the ones that are added back to the main stream are taken into the consideration for neural rejuvenation.
For DenseNet~\cite{densenet}, due to the GPU memory and speed issue, we are only able to run DenseNet with 121 layers.
We change it from pre-activation~\cite{resnetv2} to post-activation~\cite{resnet} to follow our assumption that each convolutional layer is directly followed by a batch normalization layer.
This change yields a similar accuracy to the original one.

A quick observation of our results is that the models with stronger capacities actually have better improvements from Neural Rejuvenation.
This is consistent with our discussion in Sec.~\ref{sec:lat} and the observation in Sec.~\ref{sec:res_utl}.
For large-scale tasks, the model capacity is important and larger models are more likely to waste more resources.
Therefore, rejuvenating dead neurons in large models will improve more than doing that in small models where the resources are better utilized.
In all models, DenseNet-121 is the hardest to find dead neurons, and thus has the smallest improvements.
This may explain the model compactness discussed in their paper~\cite{densenet}.
Moreover, VGG-16 with NR achieves 23.71\% top-1 error with just 21.5M parameters, far better than \cite{netslim} which achieves 36.66\% top-1 error with 23.2M.

\begin{table}[h]
\small
    \centering
    \begin{tabular}{l|cccc}
    \toprule
    Architecture & BL~\cite{morphnet} & MN~\cite{morphnet} & BL$^*$ &NR \\
    \midrule
    MobileNet-0.50 & 42.9 & 41.9 & 41.77 & 40.12 \\
    MobileNet-0.25 & 55.2 & 54.1 & 53.76 & 51.94 \\
    \bottomrule
    \end{tabular}
    \vspace{0.05in}
    \caption{Top-1 error rates of MobileNet~\cite{howard2017mobilenets} on ImageNet. The image size is 128x128 for both training and testing. The FLOPs are maintained in all the methods. BL: the baseline performances reported in ~\cite{morphnet}, MN: MorphNet~\cite{morphnet}, BL$^*$: our implementation of the baseline, and NR: Neural Rejuvenation.}
    \label{tab:mobilenet}
\end{table}

\begin{table*}[t]
\small
    \centering
    \begin{tabular}{l| ll | ll | ll}
    \toprule
    Architecture & \multicolumn{2}{c|}{Baseline} & \multicolumn{2}{c|}{Network Slimming~\cite{netslim}} & \multicolumn{2}{c}{Neural Rejuvenation}  \\
    & C10 (Params) & C100 (Params) & C10 (Params) & C100 (Params) &  C10 (Params) & C100 (Params) \\
    \midrule
    VGG-19~\cite{vggnet} & 5.44 (20.04M)& 23.11 (20.08M) & 5.06 (10.07M) & 24.92  (10.32M) & 4.19 (9.99M) & 21.53 (10.04M) \\
    ResNet-164~\cite{resnet} &	6.11 (1.70M) & 28.86 (1.73M) & 5.65 (0.94M) & 25.61  (0.96M) & 5.13 (0.88M) & 23.84 (0.92M) \\
    DenseNet-100-40~\cite{densenet} & 3.64 (8.27M) &  19.85 (8.37M) & 3.75 (4.36M) & 19.29  (4.65M) & 3.40 (4.12M) & 18.59 (4.31M) \\
    \bottomrule
    \end{tabular}
    \vspace{0.05in}
    \caption{Neural Rejuvenation for model compression on CIFAR~\cite{cifar}.
    In the experiments for ImageNet, the computational resources are kept when rejuvenating dead neurons.
    But here, we set the resource target of neural rejuvenation to the half of the original usage.
    Then, our Neural Rejuvenation becomes a model compressing method, and thus can be compared with the state-of-the-art pruning method~\cite{netslim}.
    }
    \label{tab:cifar10}
\end{table*}

Next, we show experiments on MobileNet-0.5 and 0.25 in Table~\ref{tab:mobilenet}.
They are not included in Table~\ref{tab:imagenet} because their image size is 128x128 and the learning rate follows the cosine learning rate schedule starting from 0.1~\cite{qiao2018deep}.
MobileNet is designed for platforms with low computational resources.
Our NR outperforms the previous method~\cite{morphnet} and shows very strong improvements.

\subsection{Results on CIFAR}
The experiments on CIFAR have two parts.
The first part is to use Neural Rejuvenation as a model compression method to compare with the previous state-of-the-arts when the model sizes are halved.
The results are shown in Table~\ref{tab:cifar10}.
In the second part, we show the performances in Table~\ref{tab:multiple} where we do Neural Rejuvenation for multiple times.

\vspace{-0.15in}
\paragraph{Model compression}
Table~\ref{tab:cifar10} shows the performance comparisons on CIFAR-10/100 datasets~\cite{cifar}.
CIFAR dataset is a small dataset, with 50,000 training images and 10,000 test images.
Unlike our experiments on ImageNet, here, we do not rejuvenate dead neurons to utilize all the available computational resource;
instead, we set the resource target to $0.5\times\mathcal{C}$ where $\mathcal{C}$ is the original resource constraint.
In practice, this is done by setting $T_r=0.25$ and rejuvenating the models to the level of $0.5\times\mathcal{C}$.
As a result, Neural Rejuvenation ends up training a model with only a half of the parameters, which can be compared with the previous state-of-the-art network pruning method~\cite{netslim}.

\vspace{-0.15in}
\paragraph{Multiple NR}\label{sec:multi}
Table~\ref{tab:multiple} shows the performances of VGG-19 tested on CIFAR datasets without limiting the times of Neural Rejuvenation.
The improvement trends are clear when the number of Neural Rejuvenation increases.
The relative gains are $33.5\%$ for CIFAR-10 and $13.8\%$ for CIFAR-100.
\begin{table}[h]
    \small
    \centering
    \begin{tabular}{l|cccccc}
        \toprule
        \# of NR & 0 & 1 & 2 & 3 & 4 & 5 \\
        \midrule
        C10 & 5.44 & 4.19 & 4.03 & 3.79 & 3.69 & 3.62 \\
        C100 & 23.11 & 21.53 & 20.47 & 19.91 & --- & ---\\
        \bottomrule
    \end{tabular}
    \vspace{0.05in}
    \caption{Error rates of VGG-19 on CIFAR-10 (C10) and CIFAR-100 (C100) with different times of Neural Rejuvenation while maintaining the number of parameters.}
    \label{tab:multiple}
\end{table}

Here, we introduce the detailed settings of the experiments.
For VGG-19, we make the following changes because the original architecture is not designed for CIFAR.
First, we remove all the fully-connected layers and add a global average pooling layer after the convolutional layers which is then followed by a fully-connected layer that produces the final outputs.
Then, we remove the original 4 max-pooling layers and add 2 max-pooling layers after the 4$^{\text{th}}$ and the 10$^{\text{th}}$ convolutional layers for downsampling.
These changes adapt the original architecture to CIFAR, and the baseline error rates become lower, \textit{e.g.}~from 6.66 to 5.44 on CIFAR-10 and from 28.05 to 23.11 on CIFAR-100.
We make the same changes to DenseNet as for ImageNet.
For ResNet-164 with bottleneck blocks, similar to our settings on ImageNet, we only consider the neurons that are not on the mainstream of the network for Neural Rejuvenation.
Our method is NR-CR, which removes all the cross-connections.
Table~\ref{tab:cifar10} shows that our Neural Rejuvenation can be used for training small models as well.
Table~\ref{tab:multiple} presents the potential of VGG-19 when trained with multiple times of Neural Rejuvenation.
While maintaining the number of parameters, Neural Rejuvenation improves the performances by a very large margin.

\section{Conclusion}
In this paper, we study the problem of improving the training of deep neural networks by enhancing the computational resource utilization.
This problem is motivated by two observations on deep network training, (1) more computational resources usually lead to better performances, and (2) the resource utilization of models trained by standard optimizers may be unsatisfactory.
Therefore, we study the problem of maximizing the resource utilization.
To this end, we propose a novel method named Neural Rejuvenation, which rejuvenates dead neurons during training by reallocating and reinitializing them.
Neural rejuvenation is composed of three components: resource utilization monitoring, dead neuron rejuvenation and training schemes for networks with mixed types of neurons.
These components detect the liveliness of neurons in real time, rejuvenate dead ones when needed and provide different training strategies when the networks have mixed types of neurons.
We test neural rejuvenation on the challenging datasets CIFAR and ImageNet, and show that our method can improve a variety of state-of-the-art network architectures while maintaining either their numbers of parameters or the loads of computations.
Moreover, when we target the architecture to a lower computational cost, Neural Rejuvenation can be used for model compression, which also shows better performances than the previous state-of-the-arts.
In conclusion, Neural Rejuvenation is an optimization technique with a focus on the resource utilization, which improves the training of deep neural networks by enhancing the utilization.

{\small
\bibliographystyle{ieee}
\bibliography{egbib}

\begin{thebibliography}{10}\itemsep=-1pt

\bibitem{alvarez2016learning}
J.~M. Alvarez and M.~Salzmann.
\newblock Learning the number of neurons in deep networks.
\newblock In {\em Advances in Neural Information Processing Systems}, pages
  2270--2278, 2016.

\bibitem{ba2014deep}
J.~Ba and R.~Caruana.
\newblock Do deep nets really need to be deep?
\newblock In {\em Advances in neural information processing systems}, pages
  2654--2662, 2014.

\bibitem{ba2014multiple}
J.~Ba, V.~Mnih, and K.~Kavukcuoglu.
\newblock Multiple object recognition with visual attention.
\newblock {\em arXiv preprint arXiv:1412.7755}, 2014.

\bibitem{bahdanau2014neural}
D.~Bahdanau, K.~Cho, and Y.~Bengio.
\newblock Neural machine translation by jointly learning to align and
  translate.
\newblock {\em arXiv preprint arXiv:1409.0473}, 2014.

\bibitem{baker2016designing}
B.~Baker, O.~Gupta, N.~Naik, and R.~Raskar.
\newblock Designing neural network architectures using reinforcement learning.
\newblock {\em arXiv preprint arXiv:1611.02167}, 2016.

\bibitem{cai2018efficient}
H.~Cai, T.~Chen, W.~Zhang, Y.~Yu, and J.~Wang.
\newblock Efficient architecture search by network transformation.
\newblock AAAI, 2018.

\bibitem{caron2018deep}
M.~Caron, P.~Bojanowski, A.~Joulin, and M.~Douze.
\newblock Deep clustering for unsupervised learning of visual features.
\newblock In {\em arXiv preprint arXiv:1807.05520}, 2018.

\bibitem{chen2015abc}
K.~Chen, J.~Wang, L.-C. Chen, H.~Gao, W.~Xu, and R.~Nevatia.
\newblock Abc-cnn: An attention based convolutional neural network for visual
  question answering.
\newblock {\em arXiv preprint arXiv:1511.05960}, 2015.

\bibitem{deeplab}
L.~Chen, G.~Papandreou, I.~Kokkinos, K.~Murphy, and A.~L. Yuille.
\newblock Semantic image segmentation with deep convolutional nets and fully
  connected crfs.
\newblock In {\em International Conference on Learning Representations}, 2015.

\bibitem{courbariaux2016binarized}
M.~Courbariaux, I.~Hubara, D.~Soudry, R.~El-Yaniv, and Y.~Bengio.
\newblock Binarized neural networks: Training deep neural networks with weights
  and activations constrained to+ 1 or-1.
\newblock {\em arXiv preprint arXiv:1602.02830}, 2016.

\bibitem{denton2014exploiting}
E.~L. Denton, W.~Zaremba, J.~Bruna, Y.~LeCun, and R.~Fergus.
\newblock Exploiting linear structure within convolutional networks for
  efficient evaluation.
\newblock In {\em Advances in neural information processing systems}, pages
  1269--1277, 2014.

\bibitem{NR}
Y.~Dong and E.~J. Nestler.
\newblock The neural rejuvenation hypothesis of cocaine addiction.
\newblock {\em Trends Pharmacol Sci}, 35(8):374--383, Aug 2014.

\bibitem{elsken2017simple}
T.~Elsken, J.-H. Metzen, and F.~Hutter.
\newblock Simple and efficient architecture search for convolutional neural
  networks.
\newblock {\em arXiv preprint arXiv:1711.04528}, 2017.

\bibitem{frankle2018lottery}
J.~Frankle and M.~Carbin.
\newblock The lottery ticket hypothesis: Training pruned neural networks.
\newblock {\em arXiv preprint arXiv:1803.03635}, 2018.

\bibitem{goggin1991second}
S.~D. Goggin, K.~M. Johnson, and K.~E. Gustafson.
\newblock A second-order translation, rotation and scale invariant neural
  network.
\newblock In {\em Advances in neural information processing systems}, pages
  313--319, 1991.

\bibitem{morphnet}
A.~Gordon, E.~Eban, O.~Nachum, B.~Chen, H.~Wu, T.-J. Yang, and E.~Choi.
\newblock Morphnet: Fast \& simple resource-constrained structure learning of
  deep networks.
\newblock In {\em IEEE Conference on Computer Vision and Pattern Recognition
  (CVPR)}, 2018.

\bibitem{han2016eie}
S.~Han, X.~Liu, H.~Mao, J.~Pu, A.~Pedram, M.~A. Horowitz, and W.~J. Dally.
\newblock Eie: efficient inference engine on compressed deep neural network.
\newblock In {\em Computer Architecture (ISCA), 2016}, pages 243--254. IEEE,
  2016.

\bibitem{han2015deep}
S.~Han, H.~Mao, and W.~J. Dally.
\newblock Deep compression: Compressing deep neural networks with pruning,
  trained quantization and huffman coding.
\newblock {\em arXiv preprint arXiv:1510.00149}, 2015.

\bibitem{han2015learning}
S.~Han, J.~Pool, J.~Tran, and W.~Dally.
\newblock Learning both weights and connections for efficient neural network.
\newblock In {\em Advances in neural information processing systems}, pages
  1135--1143, 2015.

\bibitem{hao2017end}
Y.~Hao, Y.~Zhang, K.~Liu, S.~He, Z.~Liu, H.~Wu, and J.~Zhao.
\newblock An end-to-end model for question answering over knowledge base with
  cross-attention combining global knowledge.
\newblock In {\em Annual Meeting of the Association for Computational
  Linguistics}, volume~1, pages 221--231, 2017.

\bibitem{hassibi1993second}
B.~Hassibi and D.~G. Stork.
\newblock Second order derivatives for network pruning: Optimal brain surgeon.
\newblock In {\em Advances in neural information processing systems}, pages
  164--171, 1993.

\bibitem{resnet}
K.~He, X.~Zhang, S.~Ren, and J.~Sun.
\newblock Deep residual learning for image recognition.
\newblock {\em IEEE Conference on Computer Vision and Pattern Recognition,
  CVPR}, 2016.

\bibitem{resnetv2}
K.~He, X.~Zhang, S.~Ren, and J.~Sun.
\newblock Identity mappings in deep residual networks.
\newblock {\em ECCV}, 2016.

\bibitem{chnprune}
Y.~He, X.~Zhang, and J.~Sun.
\newblock Channel pruning for accelerating very deep neural networks.
\newblock In {\em {IEEE} International Conference on Computer Vision, {ICCV}
  2017, Venice, Italy, October 22-29, 2017}, pages 1398--1406, 2017.

\bibitem{hinton2015distilling}
G.~Hinton, O.~Vinyals, and J.~Dean.
\newblock Distilling the knowledge in a neural network.
\newblock {\em arXiv preprint arXiv:1503.02531}, 2015.

\bibitem{howard2017mobilenets}
A.~G. Howard, M.~Zhu, B.~Chen, D.~Kalenichenko, W.~Wang, T.~Weyand,
  M.~Andreetto, and H.~Adam.
\newblock Mobilenets: Efficient convolutional neural networks for mobile vision
  applications.
\newblock {\em arXiv preprint arXiv:1704.04861}, 2017.

\bibitem{densenet}
G.~Huang, Z.~Liu, and K.~Q. Weinberger.
\newblock Densely connected convolutional networks.
\newblock {\em IEEE Conference on Computer Vision and Pattern Recognition,
  CVPR}, 2017.

\bibitem{batchnorm}
S.~Ioffe and C.~Szegedy.
\newblock Batch normalization: Accelerating deep network training by reducing
  internal covariate shift.
\newblock In {\em Proceedings of the 32nd International Conference on Machine
  Learning, {ICML}}, 2015.

\bibitem{johnson2017billion}
J.~Johnson, M.~Douze, and H.~J{\'e}gou.
\newblock Billion-scale similarity search with gpus.
\newblock {\em arXiv preprint arXiv:1702.08734}, 2017.

\bibitem{joulin2016learning}
A.~Joulin, L.~van~der Maaten, A.~Jabri, and N.~Vasilache.
\newblock Learning visual features from large weakly supervised data.
\newblock In {\em ECCV}, pages 67--84. Springer, 2016.

\bibitem{kazemy2007second}
A.~Kazemy, S.~A. Hosseini, and M.~Farrokhi.
\newblock Second order diagonal recurrent neural network.
\newblock In {\em Industrial Electronics, ISIE 2007.}, pages 251--256. IEEE,
  2007.

\bibitem{cifar}
A.~Krizhevsky and G.~Hinton.
\newblock Learning multiple layers of features from tiny images.
\newblock {\em Master's thesis, Department of Computer Science, University of
  Toronto}, 2009.

\bibitem{lebedev2014speeding}
V.~Lebedev, Y.~Ganin, M.~Rakhuba, I.~Oseledets, and V.~Lempitsky.
\newblock Speeding-up convolutional neural networks using fine-tuned
  cp-decomposition.
\newblock {\em arXiv preprint arXiv:1412.6553}, 2014.

\bibitem{lebedev2016fast}
V.~Lebedev and V.~Lempitsky.
\newblock Fast convnets using group-wise brain damage.
\newblock In {\em 2016 IEEE Conference on Computer Vision and Pattern
  Recognition (CVPR)}, pages 2554--2564. IEEE, 2016.

\bibitem{lecun1990optimal}
Y.~LeCun, J.~S. Denker, and S.~A. Solla.
\newblock Optimal brain damage.
\newblock In {\em Advances in neural information processing systems}, pages
  598--605, 1990.

\bibitem{lee2018stacked}
K.-H. Lee, X.~Chen, G.~Hua, H.~Hu, and X.~He.
\newblock Stacked cross attention for image-text matching.
\newblock {\em arXiv preprint arXiv:1803.08024}, 2018.

\bibitem{li2016pruning}
H.~Li, A.~Kadav, I.~Durdanovic, H.~Samet, and H.~P. Graf.
\newblock Pruning filters for efficient convnets.
\newblock {\em arXiv preprint arXiv:1608.08710}, 2016.

\bibitem{pnas}
C.~Liu, B.~Zoph, M.~Neumann, J.~Shlens, W.~Hua, L.~Li, L.~Fei{-}Fei, A.~L.
  Yuille, J.~Huang, and K.~Murphy.
\newblock Progressive neural architecture search.
\newblock In {\em Computer Vision - {ECCV} 2018 - 15th European Conference,
  Munich, Germany, September 8-14, 2018, Proceedings, Part {I}}, pages 19--35,
  2018.

\bibitem{netslim}
Z.~Liu, J.~Li, Z.~Shen, G.~Huang, S.~Yan, and C.~Zhang.
\newblock Learning efficient convolutional networks through network slimming.
\newblock In {\em {IEEE} International Conference on Computer Vision, {ICCV}
  2017, Venice, Italy, October 22-29, 2017}, pages 2755--2763, 2017.

\bibitem{luo2017thinet}
J.-H. Luo, J.~Wu, and W.~Lin.
\newblock Thinet: A filter level pruning method for deep neural network
  compression.
\newblock {\em arXiv preprint arXiv:1707.06342}, 2017.

\bibitem{mrnn}
J.~Mao, W.~Xu, Y.~Yang, J.~Wang, and A.~L. Yuille.
\newblock Deep captioning with multimodal recurrent neural networks (m-rnn).
\newblock {\em CoRR}, abs/1412.6632, 2014.

\bibitem{miikkulainen2019evolving}
R.~Miikkulainen, J.~Liang, E.~Meyerson, A.~Rawal, D.~Fink, O.~Francon, B.~Raju,
  H.~Shahrzad, A.~Navruzyan, N.~Duffy, et~al.
\newblock Evolving deep neural networks.
\newblock In {\em Artificial Intelligence in the Age of Neural Networks and
  Brain Computing}, pages 293--312. Elsevier, 2019.

\bibitem{mnih2014recurrent}
V.~Mnih, N.~Heess, A.~Graves, et~al.
\newblock Recurrent models of visual attention.
\newblock In {\em Advances in neural information processing systems}, pages
  2204--2212, 2014.

\bibitem{molchanov2016pruning}
P.~Molchanov, S.~Tyree, T.~Karras, T.~Aila, and J.~Kautz.
\newblock Pruning convolutional neural networks for resource efficient
  inference.
\newblock {\em arXiv preprint arXiv:1611.06440}, 2016.

\bibitem{pham2018efficient}
H.~Pham, M.~Y. Guan, B.~Zoph, Q.~V. Le, and J.~Dean.
\newblock Efficient neural architecture search via parameter sharing.
\newblock {\em arXiv preprint arXiv:1802.03268}, 2018.

\bibitem{fewshot}
S.~Qiao, C.~Liu, W.~Shen, and A.~L. Yuille.
\newblock Few-shot image recognition by predicting parameters from activations.
\newblock In {\em {IEEE} Conference on Computer Vision and Pattern Recognition,
  {CVPR}}, 2018.

\bibitem{scalenet}
S.~Qiao, W.~Shen, W.~Qiu, C.~Liu, and A.~L. Yuille.
\newblock Scalenet: Guiding object proposal generation in supermarkets and
  beyond.
\newblock In {\em 2017 {IEEE} International Conference on Computer Vision,
  {ICCV} 2017, Venice, Italy, October 22-29}, 2017.

\bibitem{qiao2018deep}
S.~Qiao, W.~Shen, Z.~Zhang, B.~Wang, and A.~Yuille.
\newblock Deep co-training for semi-supervised image recognition.
\newblock In {\em European Conference on Computer Vision}, 2018.

\bibitem{gunn}
S.~Qiao, Z.~Zhang, W.~Shen, B.~Wang, and A.~L. Yuille.
\newblock Gradually updated neural networks for large-scale image recognition.
\newblock In {\em Proceedings of the 35th International Conference on Machine
  Learning, {ICML}}, 2018.

\bibitem{rastegari2016xnor}
M.~Rastegari, V.~Ordonez, J.~Redmon, and A.~Farhadi.
\newblock Xnor-net: Imagenet classification using binary convolutional neural
  networks.
\newblock In {\em European Conference on Computer Vision}, pages 525--542.
  Springer, 2016.

\bibitem{real2017large}
E.~Real, S.~Moore, A.~Selle, S.~Saxena, Y.~L. Suematsu, J.~Tan, Q.~Le, and
  A.~Kurakin.
\newblock Large-scale evolution of image classifiers.
\newblock {\em arXiv preprint arXiv:1703.01041}, 2017.

\bibitem{ILSVRC15}
O.~Russakovsky, J.~Deng, H.~Su, J.~Krause, S.~Satheesh, S.~Ma, Z.~Huang,
  A.~Karpathy, A.~Khosla, M.~Bernstein, A.~C. Berg, and L.~Fei-Fei.
\newblock {ImageNet Large Scale Visual Recognition Challenge}.
\newblock {\em International Journal of Computer Vision (IJCV)},
  115(3):211--252, 2015.

\bibitem{vggnet}
K.~Simonyan and A.~Zisserman.
\newblock Very deep convolutional networks for large-scale image recognition.
\newblock {\em CoRR}, abs/1409.1556, 2014.

\bibitem{srivastava2015training}
R.~K. Srivastava, K.~Greff, and J.~Schmidhuber.
\newblock Training very deep networks.
\newblock In {\em Advances in neural information processing systems}, pages
  2377--2385, 2015.

\bibitem{szegedy2015going}
C.~Szegedy, W.~Liu, Y.~Jia, P.~Sermanet, S.~Reed, D.~Anguelov, D.~Erhan,
  V.~Vanhoucke, and A.~Rabinovich.
\newblock Going deeper with convolutions.
\newblock In {\em Proceedings of the IEEE conference on computer vision and
  pattern recognition}, pages 1--9, 2015.

\bibitem{sort}
Y.~Wang, L.~Xie, C.~Liu, S.~Qiao, Y.~Zhang, W.~Zhang, Q.~Tian, and A.~Yuille.
\newblock {SORT: Second-Order Response Transform for Visual Recognition}.
\newblock {\em IEEE International Conference on Computer Vision}, 2017.

\bibitem{mssar}
Y.~Wang, L.~Xie, S.~Qiao, Y.~Zhang, W.~Zhang, and A.~L. Yuille.
\newblock Multi-scale spatially-asymmetric recalibration for image
  classification.
\newblock In {\em The European Conference on Computer Vision (ECCV)}, September
  2018.

\bibitem{wen2016learning}
W.~Wen, C.~Wu, Y.~Wang, Y.~Chen, and H.~Li.
\newblock Learning structured sparsity in deep neural networks.
\newblock In {\em Advances in Neural Information Processing Systems}, pages
  2074--2082, 2016.

\bibitem{xu2016ask}
H.~Xu and K.~Saenko.
\newblock Ask, attend and answer: Exploring question-guided spatial attention
  for visual question answering.
\newblock In {\em European Conference on Computer Vision}, pages 451--466.
  Springer, 2016.

\bibitem{xu2015show}
K.~Xu, J.~Ba, R.~Kiros, K.~Cho, A.~Courville, R.~Salakhudinov, R.~Zemel, and
  Y.~Bengio.
\newblock Show, attend and tell: Neural image caption generation with visual
  attention.
\newblock In {\em International conference on machine learning}, pages
  2048--2057, 2015.

\bibitem{xu2005maximum}
L.~Xu, J.~Neufeld, B.~Larson, and D.~Schuurmans.
\newblock Maximum margin clustering.
\newblock In {\em Advances in neural information processing systems}, pages
  1537--1544, 2005.

\bibitem{yang2018knowledge}
C.~Yang, L.~Xie, S.~Qiao, and A.~Yuille.
\newblock Knowledge distillation in generations: More tolerant teachers educate
  better students.
\newblock {\em AAAI}, 2018.

\bibitem{bnprune}
J.~Ye, X.~Lu, Z.~L. Lin, and J.~Z. Wang.
\newblock Rethinking the smaller-norm-less-informative assumption in channel
  pruning of convolution layers.
\newblock {\em CoRR}, abs/1802.00124, 2018.

\bibitem{yu2017nisp}
R.~Yu, A.~Li, C.-F. Chen, J.-H. Lai, V.~I. Morariu, X.~Han, M.~Gao, C.-Y. Lin,
  and L.~S. Davis.
\newblock Nisp: Pruning networks using neuron importance score propagation.
\newblock {\em Preprint at https://arxiv. org/abs/1711.05908}, 2017.

\bibitem{des}
Z.~Zhang, S.~Qiao, C.~Xie, W.~Shen, B.~Wang, and A.~L. Yuille.
\newblock Single-shot object detection with enriched semantics.
\newblock In {\em 2018 {IEEE} Conference on Computer Vision and Pattern
  Recognition, {CVPR} 2018, Salt Lake City, UT, USA, June 18-22, 2018}, pages
  5813--5821, 2018.

\bibitem{zhong2017practical}
Z.~Zhong, J.~Yan, and C.-L. Liu.
\newblock Practical network blocks design with q-learning.
\newblock {\em arXiv preprint arXiv:1708.05552}, 2017.

\bibitem{zhou2016less}
H.~Zhou, J.~M. Alvarez, and F.~Porikli.
\newblock Less is more: Towards compact cnns.
\newblock In {\em European Conference on Computer Vision}, pages 662--677.
  Springer, 2016.

\bibitem{zoph2016neural}
B.~Zoph and Q.~V. Le.
\newblock Neural architecture search with reinforcement learning.
\newblock {\em arXiv preprint arXiv:1611.01578}, 2016.

\bibitem{zoph2017learning}
B.~Zoph, V.~Vasudevan, J.~Shlens, and Q.~V. Le.
\newblock Learning transferable architectures for scalable image recognition.
\newblock {\em arXiv preprint arXiv:1707.07012}, 2(6), 2017.

\end{thebibliography}
}

\end{document}